\documentclass{llncs}
\usepackage{times}
\usepackage{soul}
\usepackage{url}
\usepackage{hyperref}
\usepackage[utf8]{inputenc}
\usepackage{graphicx}
\usepackage{todonotes}
\usepackage{german}

\usepackage{hyperref}

\usepackage{subfigure}
\usepackage[nottoc]{tocbibind}
\usepackage{helvet}
\usepackage{courier} 
\usepackage{url}
\usepackage{prftree}

\usepackage{type1cm}         
\usepackage{amsmath}
\usepackage{makeidx}         
\usepackage{graphicx}        
\usepackage{multicol}        

\usepackage[utf8]{inputenc}
\begin{document}
\title{Künstliche Intelligenz, quo vadis?}

\author{Ulrike Barthelme{\ss}\inst{1}, Ulrich Furbach\inst{2}}
\authorrunning{U. Barthelme{\ss},  U. Furbach}
%
\institute{  \email{ubarthelmess@gmx.de}\and University of Koblenz, Germany
\email{uli@uni-koblenz.de}}
%
\maketitle              

 \bigskip

Die Geschichte der Künstlichen Intelligenz (KI) ist eine  recht wechselhafte. Von einer anfänglichen Euphorie in den 1950er Jahren ging es wellenförmig durch Höhen und Tiefen: War man zu Beginn recht optimistisch, mit logischen, symbolischen Verfahren Probleme wie maschinelles Übersetzen, Bild- und Textverstehen oder Robotersteuerung innerhalb weniger Jahrzehnte zu lösen, musste man schließlich diese anfänglichen Vorstellungen nachjustieren. Die 1980er sahen dann die Hochzeit der Expertensysteme, deren Ziel es war, Wissen von spezialisierten Experten in einem KI-System verfügbar zu machen. Danach folgte eine Periode der Ernüchterung, der sogenannte KI-Winter. Die öffentlichen Forschungsgelder wurden zurückgefahren, viele der großen Projekte, so etwa das japanische 5th Generation Project oder das europäische ESPRIT Programm wurden nicht weiterverfolgt.

Diese Haltung änderte sich allmählich, als 1997  ein  Computer den damals amtierenden Schachweltmeister besiegte und 2011   das IBM-System Watson in der Quizshow Jeopardy gegen menschliche Champions gewann. KI wird omnipräsent: 
Wir sprechen mit Siri, Alexas und ähnlichen Assistenten, die automatische Sprachübersetzung hat es zu beachtlicher Reife gebracht, und Bilder und Videos werden automatisch ausgewertet. 2017 siegt überraschenderweise  ein KI-System gegen einen Weltklasse-Go-Spieler. Die Robotik ist mittlerweile in unserem Alltag  angekommen,  sei es staubsaugend in unseren Wohnzimmern, in der Form von autonomen Fahrzeugen im Straßenverkehr und als intelligente Waffensysteme auf den Kriegsschauplätzen dieser Welt (diesen Aspekt werden wir in Abschnitt~\ref{sec:mil} gesondert diskutieren).

Mittlerweile ist KI ein riesiger Wirtschaftsfaktor. Alle großen Unternehmen be\-schäf\-tigen sich mit den Einsatzmöglichkeiten von KI und dies quer durch alle Branchen. In Deutschland wurde der Begriff \textit{Industrie 4.0} \cite{dewiki:210450663} geprägt, wodurch eine vierte industrielle Revolution angedeutet werden soll.
Kai-Fu Lee hat dies mit folgendem Zitat eindrucksvoll formuliert:
\glqq KI wird bahnbrechender als die Erfindung der Elektrizität\grqq  \cite{Lee:2018}.
 
Natürlich führt auch dieses hohe Maß an wirtschaftlicher Bedeutung zu einem breiten gesellschaftlichen Diskurs. Wir werden diesen Aspekt später in diesem Artikel aufgreifen. In den folgenden beiden Abschnitten wollen wir ein wenig hinabsteigen in die Niederungen der KI, was die Maschinerie, die Technik und Methoden angeht. Wir meinen, dass ein gewisses Grundverständnis darüber notwendig ist, um über Möglichkeiten und Grenzen in den folgenden Abschnitten zu sprechen.

\section{Maschinelles Lernen und 
Algorithmen}
\label{sec:ml}

Maschinelles Lernen gilt seit jeher als ein Teilgebiet der KI und in der Tat gibt es auch viele verschiedene Techniken und Methoden dazu. Nahezu alle oben erwähnten glänzenden KI-Erfolge beruhen auf maschinellem Lernen.
Eine Möglichkeit, die wir hier exemplarisch diskutieren wollen,  ist das \textit{überwachte Lernen}. Soll ein System z. B. lernen, Bilder  von Katzen zu erkennen, zeigt man ihm Katzenbilder als positive Beispiele, aber auch Bilder von anderen Tieren als negative Beispiele. Im Laufe dieser Lernphase werden die Komponenten des Systems so angepasst, dass die Fehlerrate beim Klassifizieren von Bildern möglichst gering wird. Die oben angesprochenen Erfolge der KI sind im wesentlichen durch  Lernverfahren erzielt worden,  die sich  an neuronalen Netzen von Lebewesen orientieren.

\begin{figure}[htbp]
\center
\includegraphics[scale=0.5]{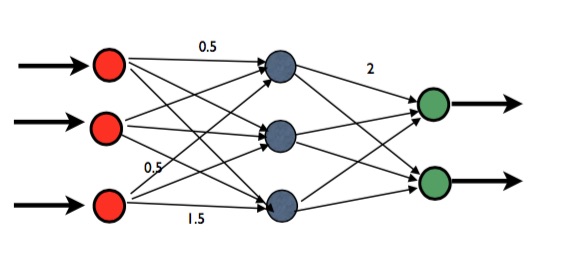} 
\caption{Künstliches neuronales Netz. An den roten Knoten liegen Eingabesignale an, die über die gewichteten Kanten weitergeleitet werden. Die grünen Knoten sind Ausgabeknoten, die schließlich die Ausgabesignale liefern. Zahlen an den Kanten bedeuten, dass die Signale entlang der Kanten mit diesem  Faktor multipliziert, also verstärkt oder abgeschwächt, werden.}
\label{fig:nn}
\end{figure}

 Vereinfacht können wir ein solches neuronales Netzwerk als Graphen, der aus Knoten und ihren Verbindungen untereinander besteht, darstellen.\footnote{die folgende Erläuterung eines neuronalen Netzes ist aus \cite{KI:Perspektiven} entnommen} Abbildung~\ref{fig:nn} zeigt ein kleines Spielbeispiel. An den roten Knoten liegen Eingabesignale an, die über die gewichteten Kanten weitergeleitet werden. 

Die grünen Knoten sind Ausgabeknoten, die schließlich die Ausgabesignale liefern. Zahlen an den Kanten bedeuten, dass die Signale entlang der Kanten mit diesem  Faktor multipliziert, also verstärkt oder abgeschwächt, werden. Was passiert nun in den einzelnen Knoten, wenn Signale eintreffen? Die Rechenkapazität eines Knotens ist denkbar einfach: Er sammelt alle eingehenden Signale auf, addiert sie, und wenn sie einen vorgegebenen Schwellwert übersteigen, gibt er das Signal weiter auf der Kante, die ihn verlässt. Nehmen wir den oberen grünen Knoten in Abbildung~\ref{fig:nn}: Drei Kanten von jeweils einem blauen Knoten erreichen ihn;  das Signal des obersten blauen Knoten wird mit 2 multipliziert und zu den Signalen der beiden anderen blauen Knoten addiert. Nehmen wir an, jeder der blauen Knoten sendet den  Wert 1; dann erhält unser grüner Knoten einmal den verstärkten Wert 2 und zweimal eine 1,  also insgesamt das  Erregungspotenzial 4. Je nach dem Schwellwert der Knoten wird nun ein Signal des grünen Ausgabeknotens gesendet. Nach diesem Schema passieren also Berechnungen mithilfe  eines solchen künstlichen neuronalen Netzes. An den Eingabeknoten liegen numerische Werte an, diese werden gemäß den Gewichten an den ausgehenden Kanten an die Knoten der nächsten Schicht weitergeleitet; in unserem Beispiel sind das die drei blauen Knoten der Mittelschicht. Dort werden die Werte aufsummiert und an die Knoten in der nächsten Schicht weitergeleitet, im Beispiel sind dies die beiden grünen Knoten, die hier auch Ausgabeknoten sind und das Ergebnis liefern.
Man kann sich nun vielleicht vorstellen, dass solche Netze dazu verwendet werden können, komplexe Berechnungen auszuführen.

\begin{figure}[htbp]
\center
\includegraphics[scale=0.4]{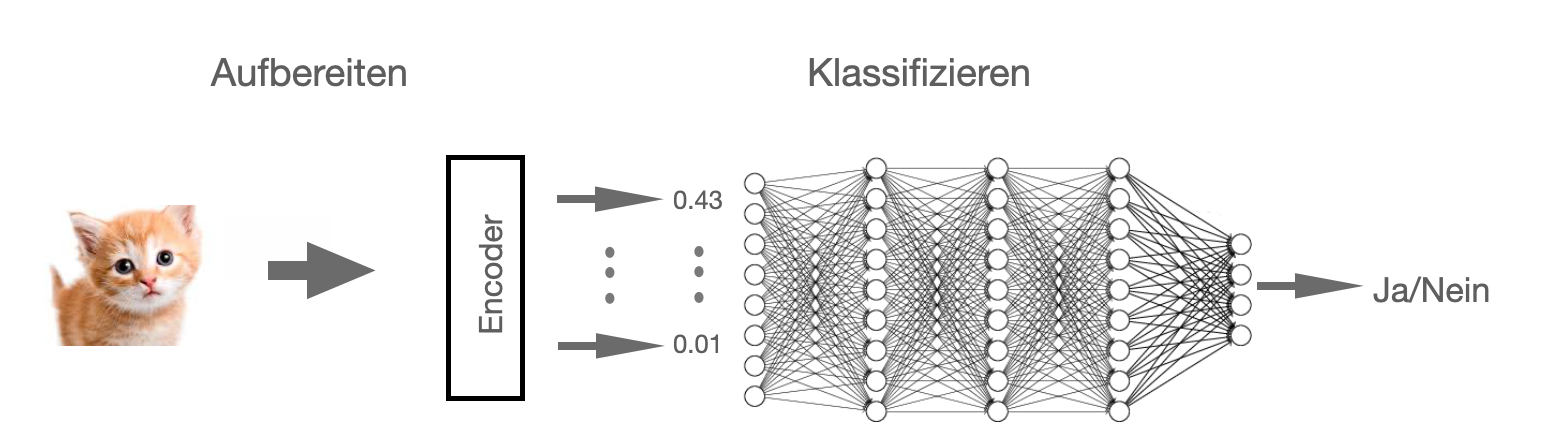} 
\caption{Katzenbild wird von einem Encoder aufbereitet und in numerische Werte zur Eingabe in ein neuronales Netz umgewandelt. Das Netz klassifiert das Bild, nachdem es trainiert wurde.}
\label{fig:prozess}
\end{figure}

Kommen wir zurück auf unser Beispiel, nämlich Katzenbilder lernen. In einem ersten Schritt muss das Bild in eine Folge von numerischen Werten konvertiert werden, da ja unsere roten Eingabeknoten Zahlenwerte erwarten. Im Prinzip können wir uns vorstellen, dass jedes Pixel des Katzenbildes einen Zahlenwert ergibt und diese nun an die Eingabeknoten des Netzwerkes übergeben werden. In  Abbildung~\ref{fig:prozess} wird dies durch das Modul \textit{Encoder} übernommen, die daran anschließende eigentliche Klassifizierung erfolgt nun ausschließlich auf der Basis der Zahlenwerte, die durch das Netzwerk \glqq geschoben\grqq  werden, wie wir es an unserem kleinen Beispielnetz in Abbildung~\ref{fig:nn} gezeigt hatten. Nun können wir auch den eigentlichen Lernvorgang erläutern: Das neuronale Netz hat initial eine gegebene Gewichtung an all seinen Kanten; wenn nun eine Katze -- bzw. das numerische Ergebnis des Aufbereitens -- präsentiert wird und das Netz klassifiziert korrekt, kann zum nächsten Bild übergegangen werden. Ist das Ergebnis jedoch falsch, also entweder erkennt das Netz eine Katze nicht oder es hält eine anderes Bild für ein Katzenbild, werden die Gewichte an den Kanten im Netz nach einem vorgegeben Verfahren leicht verändert und das Verfahren wird dann fortgesetzt. Auf diese Weise können Tausende  solcher Lernzyklen durchgeführt und das Ergebnis an neuen Bildern getestet werden, bis die Fehlerrate klein genug ist. Das Netz mit der bis dahin gelernten Gewichtung an den Kanten stellt dann ein \textit{gelerntes Modell für die Klassifikation von Katzenbildern} dar.

An dieser Stelle sollte klar sein, dass das hier demonstrierte Prinzip auf verschiedenste Weise variiert und erweitert werden kann. Es ist in den erwähnten Systemen zur Beherrschung von  Go zu finden, es wird so bei der Übersetzung von Sprache verwendet, und auch autonome Fahrzeuge analysieren auf diese Art ihre Umgebung.  In jedem Fall ist es aber so, dass die  eigentliche Berechnung -- in unserem obigen Beispiel das Klassifizieren von Bildern -- im Netzwerk durch das gewichtete Summieren in den Knoten geschieht. Dort werden lediglich Zahlenwerte berechnet; nichts gibt einen Hinweis darauf, dass wir ursprünglich das Bild eines Tieres als Ausgangspunkt hatten. Das Ergebnis bekommen wir ohne Begründung; aus der Fehlerrate des gelernten Modells bekommen wir eine Wahrscheinlichkeit für Korrektheit.

Nun liest man in vielen Diskussionen um den Einsatz der KI in kritischen Situationen, wie z.\,B.  dem Straßenverkehr, medizinischen Einsätzen oder bei Kreditvergaben, dass die verwendeten Algorithmen sicher sein und daher überprüft werden müssen. Eine umgangssprachliche Definition für Algorithmus findet sich in \cite{dewiki:210278524}:
\begin{quote}
Ein Algorithmus ist eine eindeutige Handlungsvorschrift zur Lösung eines Problems oder einer Klasse von Problemen. Algorithmen bestehen aus endlich vielen wohldefinierten Einzelschritten.
\end{quote}
Wenn wir nun ein neuronales Netz bzw. ein gelerntes Modell  für die Lösung einer Aufgabe einsetzen, läuft da natürlich ein Algorithmus ab. Die Berechnungen in dem Netzwerk, die die numerischen Eingabewerte bis zu den Ausgangsknoten weiterschieben, folgen einem Algorithmus, der in der wohldefinierten Sprache der Mathematik beschrieben ist; auch ist  die Vorschrift  eindeutig und sie besteht aus endlich vielen Einzelschritten. Wir haben dies in der Erläuterung zu Abbildung~\ref{fig:nn} deutlich gemacht. Diese einzelnen Schritte haben jedoch nichts damit zu tun, wie wir Menschen uns -- um bei unserem Katzenbeispiel zu bleiben --  davon überzeugen, dass eine Katze abgebildet ist. Wir erkennen das Fell, die Schnurrhaare, die Größe und die Gesichtsform, und vieles anderes mehr bringt uns zu der Überzeugung, dass es sich um eine Katze handelt. Der Algorithmus, den das Netzwerk ausführt, wenn es zu einer  Entscheidung kommt, manipuliert ausschließlich Zahlenwerte und hat nichts mit den Eigenschaften zu tun, die wir einer Katze zuschreiben und die wir überprüfen. Als Folge davon können wir von solch einem künstlichen neuronalen Netz auch keine Erklärung dafür erwarten, warum es sich um eine Katze handelt. Man kann sich aber durchaus Anwendungen vorstellen, in denen wir von einem KI-System erwarten, dass es uns eine Erklärung für seine Entscheidung liefert. Beispiele dafür wären  medizinische Anwendungen, KI-Systeme, die Bewerbungen in Personalabteilungen vorsortieren, oder gar militärische Anwendungen, bei denen  Menschenleben auf dem Spiel stehen. 

In der KI-Forschung gibt es daher seit einiger Zeit einen Trend hin zur erklärbaren KI. Man versucht, verschiedenartige  Techniken mit statistischen Methoden des maschinellen Lernens, wie eben neuronalen Netzen, zu kombinieren, um nicht nur Ergebnisse sondern auch Erklärungen dazu zu bekommen. Seit 2018 existiert das europäische Claire Netzwerk (Confederation of Laboratories for Artificial Intelligence Research in Europe)\footnote{\url{https://claire-ai.org/}, aufgerufen am 19.4.2021}, das sich auch der Vision verschrieben hat, vertrauenswürdige KI-Systeme zu entwickeln, die die menschliche Intelligenz ergänzen, anstatt sie zu ersetzen, und die somit den Menschen zugutekommen.

\section{Das Problem mit dem Wissen}
\label{sec:wissen}
Das  bisher beschriebene Vorgehen lässt sich auch unter dem Schlagwort \textit{statistische Methoden} einordnen. Man benutzt sehr viele Daten, um zu lernen. Im Gegensatz dazu gibt es eine  reiche Tradition an KI-Methoden, die symbolisches Wissen einsetzen, um Probleme zu lösen. Hierbei wird Wissen mithilfe einer formalisierten Sprache so dargestellt, dass es von Computern verarbeitet werden kann. Im  folgenden Beispiel wird Wissen durch logische Formeln zusammen mit einer Schlussfolgerung dargestellt. Über dem waagerechten Strich stehen Prämissen und darunter die Konklusion:

\begin{displaymath}
\prftree
{\prfassumption{\forall x (mensch(x) \rightarrow sterblich(x))}   
\hskip 5mm
\prfassumption{ mensch(sokrates)} 
}
{{sterblich(sokrates)}}
\end{displaymath}

Das Symbol $\forall$ steht hierbei für \textit{für alle} und  $\rightarrow$ für \textit{impliziert}. Die erste Prämisse liest sich damit \textit{für alle Objekte x gilt, wenn x ein Mensch ist, ist x auch sterblich}.	Mithilfe der Schlussregel kann damit aus den beiden Prämissen das neue Wissen, nämlich dass Sokrates sterblich ist, hergeleitet werden.
Deutlich wird durch dieses Beispiel auch, dass es sich hier um die Formulierung von Wissen über die Welt  und nicht um mathematische Sachverhalte handelt. Wir können uns vorstellen, dass ein System, das solche Mechanismen verwendet, auch Erklärungen liefern kann. Im Beispiel etwa \textit{Sokrates ist sterblich, weil er ein Mensch ist!}

Nun könnte man argumentieren, dass man lediglich mit solchen symbolischen Verfahren arbeiten müsste, um zu erklärbarer KI zu kommen. Dass dies leider nicht so einfach ist, wollen wir anhand des Alltagsschließens (commonsense reasoning), einem etablierten und wichtigen   Teilgebiet der KI, verdeutlichen. Nehmen wir folgendes Beispiel aus einer Test-Datenbank für Alltagsschließen: Die Aufgabe ist dabei, einen Sachverhalt zu erklären, z. B. die Aussage \textit{Mein Körper wirft einen Schatten auf das Gras.}.  Die Frage ist nun \textit{Was war die Ursache dafür?} und die Test-Datenbank enthält dafür zwei  alternative Antworten: \textit{Die Sonne ist aufgegangen.} und\textit{ Das Gras wurde gemäht.} Das System muss nun die plausiblere der beiden Antworten auswählen. Man macht sich leicht klar, welches Wissen notwendig ist, um solche für den Menschen triviale Fragen zu beantworten. Man muss wissen, dass Schatten durch Beleuchtung zustande kommen und die aufgehende Sonne eine Lichtquelle darstellt, während die Beschaffenheit des Grases wenig mit dem Schattenwurf zu tun hat. Das Problem für ein künstliches System ist dabei, diese relevanten Teile des Wissens in der riesigen Wissensbasis zu identifizieren und dann auch zu verarbeiten. Das Verarbeiten selbst, also das logische Schlussfolgern aus einzelnen Fakten, ist bereits sehr gut untersucht. Es existieren zahlreiche leistungsfähige Systeme für logikbasiertes automatisches Schließen, die auch in der industriellen Software-Entwicklung eine wichtige Rolle spielen. 

Nun findet man mittlerweile auch  umfangreiche Wissensquellen, die man mit solchen wissensbasierten Systemen kombinieren kann. So  benutzt, pflegt und erweitert der Suchmaschinen-Betreiber Google die sogenannten Knowledge Graphs, die Wissen in formalisierter Form zur Verbesserung von Suchanfragen enthalten. Im Mai 2020 waren darin 500 Milliarden Fakten zu 5 Milliarden Begriffen gespeichert. 
Trotz dieser riesigen Menge an Daten finden sich Suchmaschinen in Sekundenbruchteilen darin zurecht. Die Situation ist jedoch ungleich komplexer, wenn nicht nur etwas gesucht wird, sondern wenn Fakten-Wissen dazu benutzt werden soll, um Zusammenhänge, Erklärungen oder Begründungen zu finden. Wenn man also solche  Wissensbasen in einem logikbasierten  Schlussfolgerungssystem verwenden möchte, stößt man sehr schnell an Grenzen der Komplexität. Und so kommt es, dass im Bereich des Alltagsschließens statistische Methoden deutlich besser sind (\cite{kocijan2020review}), mit dem offensichtlichen Nachteil, dass sie nicht auch unmittelbar Erklärungen liefern können.

 Im Forschungsprojekt CoRg  (Cognitive Reasoning, http://corg.hs-harz.de) wird derzeit ein System entwickelt, das versucht einen wissensbasierten logischen Ansatz mit statistischen Methoden zu kombinieren.

\section{Moralische Maschinen}

Wir haben oben angesprochen, dass KI-Systeme für autonomes Fahren bereits erfolgreich entwickelt und auch in bestimmten Bereichen eingesetzt werden. Natürlich be\-schäftigt sich auch die  Politik mit  Regulierungsmöglichkeiten  für autonome Fahrzeuge im öffentlichen  Straßenverkehr. Die Frage ist hier, neben der allgemeinen Sicherheit, ob und wie ein autonomes Fahrzeug das sogenannte Weichenstellerproblem (Trolley Problem) lösen wird. Hierbei handelt es sich um ein ethisch moralisches Dilemma: 
\begin{quote}
Sie befinden sich am Weichenstellmechanismus  einer Gleisanlage. Ein Zug rast auf eine Weiche zu, er droht dabei fünf Personen zu überrollen. Wenn Sie die Weiche betätigen, fährt der Zug auf das andere Gleis, wo er dann aber eine Person überrollt. 
\end{quote}

Wie soll sich der Weichensteller entscheiden? Wenn er nicht handelt, werden fünf Menschen getötet, wenn er handelt und die Weiche umstellt, wird eine Person getötet. Solche oder ähnliche dilemmatischen Situationen kommen in verschiedensten Bereichen vor, etwa in der Medizin oder der Rechtsprechung. 
Diskutiert werden solche Fragestellungen seit Langem in der Philosophie und dabei haben sich zwei gegensätzliche Auffassungen herauskristallisiert: eine utilitaristische Auffassung, nach der es vertretbar ist, ein Leben zu opfern, um fünf zu retten, und eine ethische und normenorientierte Auffassung, nach der die Norm, niemanden zu verletzen (also nicht handeln), als stärker angesehen wird als die Norm, andere zu retten (also die Weiche umstellen). Welches Vorgehen soll nun in einem autonomen Fahrzeug in einer dem Weichenstellerproblem vergleichbaren Situation programmiert werden? Der Bundesverkehrsminister Deutschlands hat eine Ethikkommission eingesetzt, die über automatisiertes und vernetztes Fahren im Juni 2017 einen Bericht vorgelegt hat. Darin wird ganz klar formuliert: \glqq Technische Systeme müssen auf Unfallvermeidung ausgelegt werden, sind aber auf eine komplexe oder intuitive Unfallfolgenabschätzung nicht so normierbar, dass sie die Entscheidung eines sittlich urteilsfähigen, verantwortlichen Fahrzeugführers ersetzen oder vorwegnehmen könnten\grqq   \cite{ethik-kommisssion}. Tatsächlich scheint es derzeit nicht machbar, ethische Normen so zu formulieren und damit auch zu fixieren, dass sie von einem KI-System in komplexen und verschiedenartigsten Situationen angewendet werden können. 

Wie komplex die Fragestellungen im Zusammenhang mit autonomen Systemen sind, zeigt auch ein anderer Teil des Berichts der Ethikkommission, nämlich wenn das Thema 'selbstlernende Systeme' behandelt wird. Diesem Bericht wird  eine lange Liste von Forderungen, die von autonomen und vernetzten Fahrzeugen erfüllt werden sollen, vorangestellt. Wobei gerade Fragen der Überprüfbarkeit der Systeme im Vordergrund stehen.


Wenn nun aber ein künstliches System lernt und sich damit weiterentwickelt, könnte es ja durchaus vorkommen, dass es sich dann im Vergleich zu seinem ursprünglichen Zustand, also dem vor dem Selbstlernen, anders verhält. Wir haben im Abschnitt~\ref{sec:ml} darauf hingewiesen, wie schwierig es ist, Entscheidungen von Systemen, die auf neuronalen Netzen basieren, zu erklären oder zu überprüfen. 
Das Fahrzeug würde dann ein Verhalten an den Tag legen,  das sich keiner erklären kann. 

Die Ethikkommission umgeht dieses Problem, indem sie erklärt: \glqq Ein Einsatz von selbstlernenden Systemen ist beim gegenwärtigen Stand der Technik daher nur bei nicht unmittelbar sicherheitsrelevanten Funktionen denkbar.\grqq


\section{Bewusste Maschinen}

Das Weichensteller-Problem führt uns deutlich vor Augen, mit welchen Fragestellungen die zukünftige Entwicklung der erklärbaren KI behaftet ist. Fragestellungen, denen auch die Menschen nicht wirklich gewachsen sind.

Dennoch sollte man nicht aus den Augen verlieren, dass der Umgang mit der enormen Datenmenge, die wir bei der Bewältigung von Aufgaben heranziehen können, effizienter wäre, wenn KI 
nicht nur statistische Operationen (Auffinden und Abgleichen von Daten, wiederholendes Training) sondern auch kreative Funktionen (Assoziationen, Alltagsschließen, Gedankenwandern) ausüben könnte.

Wir greifen auf das obengenannte Beispiel zurück: 
\textit{Mein Körper wirft einen Schatten auf das Gras.} Ein Mensch wird, ohne zu zögern, die richtige Ursache für das Phänomen nennen können. Warum? Er hat eine lebenslange Erfahrung mit seinem Körper in der Welt und weiß, dass die Sonne als Lichtquelle für seinen Schatten zuständig ist. Er hat beobachtet, dass der Schatten sich bewegt, wenn er sich bewegt, dass der Schatten zu ihm gehört, dass Licht und Schatten miteinander in Verbindung stehen. Der Schatten ist Teil seiner selbst und dessen ist er sich bewusst. 

Kann diese Form von Bewusstsein auf KI-Systeme übertragen werden?
Als nicht-biologische Geschöpfe verfügen sie zwar nicht über phänomenales Bewusstsein wie Menschen oder andere Lebewesen, dennoch könnte man 
ihnen einen gewissen Grad an Bewusstheit zubilligen, wenn man sich auf die 
Perspektive des \textit{Graduellen Panpsychismus} einlässt.
Dieser von Patrick Spät eingeführte Begriff geht von einer fundamentalen Verbindung zwischen Geist und Materie aus. Dies bedeutet, dass nicht nur Menschen und Tiere, sondern auch Zellen, Bakterien und sogar Elektronen zumindest rudimentäre mentale Eigenschaften besitzen. Dabei ist die einfachste  rudimentäre Form von Geist Fähigkeit zur Informationsverarbeitung. 

Und genau diese Fähigkeit wird auch von machen Neurowissenschaftlern als Grundlage einer Definition des Bewusstseins benutzt.    
Die Information Integration Theory von Tononi \cite{tononi:04} z. B. geht von einem beliebigen vernetzten physischen (nicht notwendigerweise biologischen) System aus. Mit Hilfe einer mathematischen Strukturanalyse wird ermittelt, in welchem Maße das System Informationen integrieren kann. Dabei soll identifiziert werden, welche Teile des Systems Informationen in einem räumlich-zeitlichen Rahmen integrieren, um einen kohärenten und einheitlichen Zustand zu erreichen, der nicht weiter auf elementare Bestandteile reduziert werden kann. Dieser Bereich des Systems stellt in dieser Theorie den Sitz der Qualia dar. Die ermittelte Maßzahl bezeichnet auch den Grad des Bewusstseins, über den das System verfügt. 
Eine weitere Theorie, die Global Workspace Theory von Baars \cite{Baars:97}, auch als Baars’ Theater  bekannt, betont,  dass Bewusstsein unabdingbar ist, um große Mengen an Wissen zu beherrschen. Die Theater-Metapher beschreibt dabei eine Bühne mit Scheinwerfer (Aufmerksamkeit), auf der die Akteure darum eifern, in das Licht des Scheinwerfers zu gelangen. Hinter der Bühne, also im Dunkel des Unbewussten, ist eine Menge an Leuten aktiv (Techniker, Autoren, Regie), um das Geschehen zu unterstützen. Im Zuschauerraum befindet sich eine große Anzahl an Menschen, die Wissen, Fertigkeiten und Erfahrungen repräsentieren. Dieses Gesamtsystem ist das Bewusstsein! Zusammen mit den Überlegungen von Tononi wird klar, dass das Phänomen Bewusstsein keine Alles-oder-nichts-Eigenschaft ist, sondern als graduell vorhanden anzusehen ist.

Wir hatten im Abschnitt~\ref{sec:wissen} über Wissensverarbeitung  die Probleme beim Umgang mit sehr großen Wissensmengen angesprochen. Nimmt man sich die  Fähigkeit des Menschen, die riesige Menge an Wissen, Erfahrungen und Erinnerungen, über die er verfügt und die er (zumeist) auch mühelos abrufen kann, als Vorbild, so kann man durchaus auch KI-Systeme mit Bewusstsein in Zusammenhang bringen. Man versucht Mechanismen, wie sie in der Global Workspace Theory beschrieben sind, einzusetzen, um Wissen zu benutzen und zu verwalten. Das System verfügt damit im oben beschriebenen Sinn über Bewusstsein.

\section{Singularität oder Partner}

Wir haben nun bereits über KI in Zusammenhang mit  Lernen, mit Moral und Ethik und mit Bewusstsein gesprochen. Man könnte auf den Gedanken kommen, dass KI  immer menschenähnlicher wird und wir als Krone der Schöpfung Konkurrenz bekommen. Müssen wir davor Angst haben? Ist das eine Bedrohung? In der Tat gibt es durchaus ernstzunehmende Forscher, die eine Singularität voraussagen.
Damit wird   ein Wendepunkt bezeichnet, an dem das Zusammenspiel von Menschen und künstlicher Intelligenz so fortgeschritten ist, dass sich eine Superintelligenz heranbildet, die sich selbst weiterentwickelt und von uns Menschen nicht mehr kontrolliert werden kann. Ein prominenter Vertreter dieser These ist Ray Kurzweil. In seinem Bestseller \cite{kurzweil2005singularity} analysiert  der renommierte Informatiker, Erfinder und Unternehmer  die Entwicklung der  KI-Forschung und er prognostiziert, dass es bis zum Jahr 2029 möglich sein wird, das gesamte menschliche Gehirn 
in einem digitalen Computer zu emulieren. Solche Systeme könnten sodann analysiert und so weiterentwickelt werden, dass sie sich  bis zum Jahr 2045  radikal selbst modifiziert und weiterentwickelt haben, sodass die Singularität stattfinden kann. Diese Superintelligenz kann sich dann von unserem Planeten ausgehend verbreiten, bis sie das gesamte Universum einnimmt. Das klingt nach moderner Science-Fiction, hat aber durchaus Wurzeln in der Philosophie und sogar in der Theologie. Gerade der Aspekt, dass die Menschheit  mit dem Universum eins wird, erinnert in verblüffender Weise an die Lehren von Pierre Teilhard  de Chardin. Dieser Jesuit, Theologe und Naturwissenschaftler hatte schon am Anfang des 20. Jahrhunderts über die Weiterentwicklung des Menschen geschrieben. Seine Schriften wurden vom Vatikan abgelehnt, und erst nach seinem Tod im Jahre 1955 wurden sie veröffentlicht und erfuhren starke Beachtung. In seinem zentralen Buch \glqq Der Mensch im Kosmos\grqq  \cite{de2005mensch} beschreibt de Chardin, dass die Menschheit und das Universum sich weiterentwickeln und auf einen Endpunkt, den \glqq Punkt Omega\grqq, zubewegen, an dem Mensch, Universum und Gott eins werden. Diese Sichtweise wird auch von zeitgenössischen Kosmologen aufgegriffen und nun sogar von manchen  KI-Forschern.

Nach unserer Einschätzung sieht die Mehrzahl der KI-Forscher das Ziel ihrer Disziplin unter dem Schlagwort \textit{KI für den Menschen}. KI-Techniken und KI-Methoden werden entwickelt, um den Menschen zu unterstützen, um mit Menschen zu kooperieren. Und in der Tat können wir solches täglich beobachten, z. B. wenn wir die Übersetzungsfunktion unseres Smartphones benutzen, mit Siri oder Alexa  kommunizieren oder wenn das Fahrassistenz-System unseres Autos aktiv ist. Dabei sollte aber durchaus klar sein, dass die Verbreitung von KI-Systemen auch die Arbeitswelt verändert und dabei auch Tätigkeiten, die bislang Menschen vorbehalten waren, von KI-Systemen übernommen werden. Hier ist es  sinnvoll, einen breiten interdisziplinären und gesellschaftlichen Diskurs um die Gestaltung der Arbeitswelt anzustoßen  und diesen auch kontinuierlich weiterzuführen, um auf neue Entwicklungen reagieren zu können.

\section{Verzerrungen}

Die jüngsten Erfolge der KI beruhen zumeist auf maschinellen Lernverfahren mittels  künstlichen neuronalen Netzen. Wir haben diskutiert, dass zum Trainieren dieser Verfahren sehr viele Daten notwendig sind. Hier hat die technologische Entwicklung sehr stark mitgeholfen, da in den vergangenen Jahren immer mehr Daten öffentlich verfügbar wurden und somit auch für das Trainieren der KI-Systeme zur Verfügung stehen. Es ist ein Leichtes, Bilder, Videos oder auch Texte mit wenigen Mausklicks von verschiedensten Quellen im Internet herunterzuladen, um damit KI-Systeme zu trainieren. Dabei muss uns allerdings klar sein, dass nur das gelernt wird, was in den Daten enthalten ist. Wenn zum Beispiel Videos verwendet werden, in denen Gewalttätigkeit öfter mit Schwarzen als mit Weißen in Verbindung gebracht wird, wird das KI-System diese Verzerrung (engl. bias) eben auch lernen und dann im Einsatz entsprechend verzerrte Ergebnisse liefern. In   \cite{BOLD}
 wird dieses Problem anhand texterzeugender Systeme untersucht.  GPT-3 ist ein Beispiel für solch einen  Textgenerator, der mit sehr vielen Texten trainiert wurde und dann ein beliebiges Stück Text weiterschreiben kann, so dass nicht oder nur sehr schwer zu erkennen ist, dass es kein Mensch ist, der ihn verfasst hat \cite{brown2020language,enwiki:1015703033}. Wenn nun die Texte, mit denen GPT-3 trainiert wurden, bestimmte gesellschaftliche Vorurteile ausdrücken, wie z. B. gegenüber dem Islam oder gegenüber Frauen, werden diese sich dann auch in den erzeugten Texten wiederfinden. Das System übernimmt also Ungerechtigkeiten, die es in der Welt bereits gibt. Die Forschungsfrage, die in \cite{BOLD} untersucht wird, ist nun, solche Verzerrungen durch bestimmte Testdaten zu entdecken. Dann kann auch versucht werden, solchen Ungerechtigkeiten entgegenzuwirken.

\section{Militärische Anwendungen}
\label{sec:mil}

Die Idee, KI auch in Waffensystemen einzusetzen, ist naheliegend und mag auch manchem wünschenswert erscheinen. So könnte man als Argument anführen, dass Menschenleben verschont würden, wenn man Roboter anstatt menschliche Kombattanten auf das Gefechtsfeld schickt. Andererseits kann man aber beobachten, dass die Bereitschaft, sich auf bewaffnete Konflikte einzulassen, sehr viel höher ist, wenn Menschen nicht unmittelbar an den Kampfhandlungen beteiligt sind. So nimmt man zum Beispiel die große Anzahl an Menschen, die in verschiedensten Teilen der Welt durch Drohneneinsätze getötet oder verletzt werden, stillschweigend hin, während die Resonanz möglicherweise anders ausfiele, wenn bemannte Kampfflugzeuge eingesetzt würden. Diese Drohneneinsätze gegen Personen, die sich in Staaten aufhalten, in denen der Einsatz von Streitkräften nicht möglich ist, werden auch als \textit{gezielte Tötungen} bezeichnet und verstoßen klar gegen Völkerrecht. Diese Problematik wird in verschiedenen Kontexten diskutiert; so hat sich z. B. auch der Deutsche Bundestag mehrfach mit der Problematik der gezielten Tötungen beschäftigt, insbesondere im Zusammenhang mit der Tatsache, dass die US Air Base in  Ramstein mit Hilfe einer Satellitenstation für die US-Drohnenkriegführung unverzichtbar ist.
 
Viele KI-Wissenschaftler und Organisationen haben sich mittlerweile einer internationalen Initiative angeschlossen, die sich gegen den Einsatz von tödlichen autonomen Waffen einsetzt\footnote{\url{https://autonomousweapons.org}}. Eine weitere Initiative engagiert sich dafür, die Gefahr eines \glqq Krieges aus Versehen\grqq { }ernst zu nehmen und Entwicklungen, die dazu führen können, zu verhindern \footnote{\url{https://atomkrieg-aus-versehen.de}}. Insbesondere KI-basierte Entscheidungssysteme sind nur schwer von Menschen zu kontrollieren und auf Grund sehr geringer Reaktionszeiten auch nur schwer zu korrigieren. 

Das Problem ist dabei auch, dass mitunter nicht leicht zu erkennen ist, inwieweit KI-Techniken in Waffensystemen zur Anwendung kommen und inwieweit sie sich noch von Menschen kontrollieren lassen. So hat z. B. die US Luftwaffe 2020 erstmals ein KI-System als Co-Piloten in einem U-2 Aufklärungsflugzeug eingesetzt. Während des Fluges steuerte der KI-Algorithmus Sensoren und taktische Navigationssysteme des Jets. Wie genau die Aufgabenteilung zwischen KI und Mensch aussieht, ist unklar.

\section{Schlussbemerkungen}

Wir haben versucht, die grundlegenden Techniken für maschinelles Lernen und  Wissensverarbeitung aufzuzeigen. Wir haben dabei auch einige Fragen bezüglich der Ethik und des Bewusstseins von KI-Systemen angesprochen und anschließend einige problematische Aspekte des Einsatzes von KI-Systemen diskutiert. Die interessierte Leserin findet sehr viel Ausführlicheres darüber in \cite{KI:Perspektiven}. In diesem Beitrag sollte klar geworden sein, dass KI nicht per se gut oder schlecht ist, es kommt darauf an, wie wir sie nutzen. Wir sollten uns der Grenzen und Gefahren bewusst sein und diese auch in einem ständigen gesellschaftlichen Diskurs ausloten. Ein schönes Beispiel dafür ist die Studie \cite{Fra:2020} der \textit{Agentur für Grundrechte der Europäischen Union}, welche genau dieses fordert und dies besonders auf europäischer Ebene einfordert.
 \bibliographystyle{plain}
\bibliography{wissen}

\end{document}